\documentclass[letterpaper]{article}

\usepackage{natbib,alifeconf}  
\usepackage{hyperref}
\usepackage{balance}

\title{Artificial Life in Game Mods for Intuitive Evolution Education}
\author{Anya E. Vostinar$^{1}$, Barbara Z. Johnson$^{2}$, \and Kevin Connors$^2$ \\
\mbox{}\\
$^1$Carleton College, Northfield, MN 55057 \\
$^2$Grinnell College, Grinnell, IA 50112 \\
vostinar@carleton.edu} 
\begin{document}
\maketitle

\begin{abstract}
The understanding and acceptance of evolution by natural selection has become a difficult issue in many parts of the world, particularly the United States of America. The use of games to improve intuition about evolution via natural selection is promising but can be challenging. We propose the use of modifications to commercial games using artificial life techniques to `stealth teach' about evolution via natural selection, provide a proof-of-concept mod of the game Stardew Valley, and report on its initial reception.
\end{abstract}

\section{Introduction}

Teaching the theory of evolution via natural selection has become a controversial issue in the United States educational system in recent decades \citep{sinatra2008changing}. Even after setting aside the political and religious motivations for resistance, there remains a central challenge that evolution via natural selection is counter-intuitive partially because it occurs at spatial or temporal scales that are difficult or impossible for humans to observe, let alone understand. There are many efforts underway to improve this problem of observable evolutionary dynamics, and one of them is to incorporate evolution into games. This approach enables people to observe evolution in a context that can be non-confrontational and on temporal and spatial scales that they can grasp.

There have been a number of games that attempt to incorporate evolution as a central or secondary game mechanic, however they run into several recurring issues. One such issue is that evolution via natural selection is an inherently directionless process and there is a great temptation by game developers to add direction to it to make a game more fun. This change, however, reinforces a common misconception of evolution via natural selection, namely that of some being (either a deity or the organisms themselves) directing the process towards a goal (e.g. Spore \citep{bean2010spore}). Some educational games have a primary aim of demonstrating evolution by natural selection correctly, but end up being games that are not particularly fun to play. Even when a developer succeeds at creating a game that is both scientifically accurate and fun, they face an uphill battle to gain a large fanbase and achieve the goal of demonstrating evolution via natural selection to a large audience. Finally, if a game markets itself as being a game about evolution via natural selection, those who are resistant to accepting the idea of evolution are unlikely to even try the game. Therefore, if the goal is to use games to improve evolution understanding and acceptance, a game would have to 1) correctly implement evolution via natural selection, 2) be fun, 3) amass a large fanbase, and 4) not be overtly about evolution via natural selection. 

Creating a new game that will achieve those four criteria is a difficult problem, however we suggest an alternative. Some games support community modifications of the game code, called mods, that allow programmers to tack on extra features to an existing commercial-off-the-shelf (COTS) game. By modding existing commercial games, researchers ensure evolution is correctly implemented, are able to use a fun game with a large fanbase (assuming the game is already commercially successful), and could even utilize `stealth-learning' by pitching the mod as making the game more realistic \citep{sharp2012stealth}. Some studies have begun to explore using this approach in both formal school settings where teachers create educational mods of popular games \citep{moshirnia2007educational} and during out-of-school learning opportunities such as after school coding clubs in which learners modify COTS games to develop or demonstrate their understanding of academic content \citep{bertozzi2014using, hayes2009not, steinkuehler2009computational}.

Here we discuss how research in artificial life could be harnessed for educational goals involving evolution via natural selection through games and present a proof-of-concept mod to demonstrate the possibility of this approach. We developed this mod for the popular indie game Stardew Valley.  This game, released on February 26th, 2016, is a ``country-life RPG'' that allows interactions with plants and animals through farming, breeding, and fishing \citep{stardew_valley}.  This theme, combined with the fact that over 5 million people own the game \citep{valve_leak} and that there already exists an active modding community (of both programmers producing mods and players downloading mods) makes Stardew Valley the perfect avenue for this proof-of-concept mod. The developed mod targets the fishing feature of the game, adds functionality and possibly fun to that feature, and demonstrates a way that evolution via natural selection can be integrated into existing games.

\section{Background}

Games and virtual worlds have been used and tested in formal and informal educational settings to teach a variety of subjects, including STEM, as long as the technology has existed. For many educators, particularly in K-12 classrooms, games (either physical or digital) are seen as a way to engage and motivate the new generation of learners who prefer and are increasingly adept at learning concepts and complex tasks through a variety of media  \citep{sharp2012stealth, dahleen2014towards}.  Since Net Gen students (also called ``Digital Natives'') have grown up surrounded by digital media and routinely play computer (or smartphone) games even as young children, the traditional methods of direct instruction are ineffective tools for 21st century learners who thrive by learning through multimedia-supported exploration and knowledge construction \citep{prensky2010teaching}.  

It is important to note that, in this context, the term “games” includes a wide variety of ludic tools.  Games may include those played with physical manipulatives, such as board and card games, as well as digital games played on a computer or tablet device - with or without Internet connection.  The games used may also be either those designed for instruction (called “serious games” or “serious educational games” aka “SEGs”) or  commercial games designed for entertainment but repurposed as educational tools.  A number of studies have demonstrated success in using serious or commercial games, either physical or digital, in both the classroom and in informal situations such as afterschool programs.  However, attention needs to be paid to both the questions of whether or not the game is fun enough to hold the attention of the students \citep{baker2008trying} and also to what games are appropriate to particular pedagogical approaches and levels of learning.  Studies have shown that games (physical and digital) can be used to support traditional, instructionalist paradigms as well as a variety of active learning strategies designed to engage students in higher order thinking, reaching into the highest level of Bloom’s taxonomy: creating \citep{krathwohl2009taxonomy}.  

\subsection{Direct instruction and practice games}

Early educational games were often designed within the traditional, direct instruction framework in which the computer substituted for the teacher in providing information or demonstrating a skill.  These edutainment ``games'' were often digital, interactive versions of quizzes, flashcards, and automatically scored worksheets and were not particularly fun or game-like in themselves \citep{bertozzi2014using}.  As an early attempt to mimic the stereotypical classroom, these games generally targeted lower-order learning objectives in which students were expected to remember specific facts without necessarily being able to apply them to novel situations \citep{krathwohl2009taxonomy}.   

\subsection{Experiential or inquiry-based learning}

While direct instruction games focused on aiding students in remembering specific facts, active learning games were early experiments in creating a learning environment in which students would gain virtual experiences and collect information about a system in order to form and test hypotheses about what is happening in the system.  Students could be guided by an instructor or the in-game instructions to run a simulation of a natural system and record results \citep{christensen2009evolution, gibson2015red, jordens2018interrelating} in order to understand how that system works and propose hypotheses that explain results or predict future events as they continue to play the game.  

In other science learning games such as Quest Atlantis \citep{barab2005making}, River City \citep{clarke2009design}, and Citizen Science \citep{gaydos2012role}, students take on the role of scientists to investigate a virtual world problem such as lake pollution.  As students would maneuver their game avatar through a virtual world they would make decisions about where and how to gain information in order to form hypotheses about problems being experienced and to propose solutions.  Often these learning objectives were posed as quests, a popular fantasy game trope, that provides players with a defined, achievable goal that can be reached by a variety of paths. This increases a sense of agency and empowerment in the player/learner, and the intrinsic reward of solving the puzzle or fixing the game-world problem keeps learners motivated to overcome challenges and increase learning or skills along the way.  

\subsection{Creative platforms for learning and assessment}
Often, educators carry the burden of finding or creating a game to use in teaching their course content, however, it is possible to pass over that responsibility to the learners and challenge them to design a game while they increase mastery of academic content.  Game creation or modding has been used as a creative platform with a specified rule/tool set that guides or inspires their development of understanding of educational content - particularly of large systems \citep{dahleen2014towards}.  Such constructivist learning approaches were described by \cite{lave1991situated} and are being tested as a means for learning academic content such as second languages \citep{monterrat2012learning}.  

Games and game development processes can also be used by students as creative platforms to demonstrate their understanding of concepts that they developed by creating new games or modifying existing ones. For instance, students who mastered fractions in an elementary math program were asked to create a game to teach those concepts to younger students \citep{shreve2005let}, demonstrating proficiency as well as creating an instructional tool that the teacher could later use with new students.    

\subsection{Stealth learning or persuasive games}
Finally, students can learn by using non-traditional activities, such as playing a game without overt learning content that nevertheless teaches desired skills and academic content.  Called “stealth learning”, this teaching technique is a way to engage students in a learning activity without their awareness that their fun has a serious purpose.  For instance, entertainment-focused board and computer/video games have been used in the classroom to teach a wide variety of academic skills and content \citep{shreve2005let, sharp2012stealth, squire2004replaying}.  

Stealth learning and persuasive games pose a unique challenge since their purpose is to teach without seeming to teach.  Hence, they cannot be overtly serious or educational in nature.  They also need to be fun and have sufficiently high quality graphics and challenging game play to entice players to pick up and continue to play the game for enjoyment, hopefully recruiting friends to also play the game so that the education or persuasive messages will spread without a teacher imposing extrinsic motivation in the form of grades.  This requirement to make an educational game look and play like a commercial, entertainment-oriented game is a serious challenge since few educators have the skills or resources to create and distribute them.  Hence, there is increasing interest in modding games, either through modding commercial games or by using serious games that have proved to be successful in engaging students to keep playing outside of the classroom.  

Studies have primarily concentrated on these approaches using games/virtual worlds specifically designed for educational purposes, but far fewer studies have concentrated on using a widely-used commercial game, designed for entertainment, as the basis for science education, particularly in biology and evolution.  Squire did a study with students where they successfully learned about history through playing Civilization III \citep{squire2004replaying}, but the video game Spore has been problematic when used to teach about evolution.  In particular, Spore highlights how entertainment features of a game may lead to students developing misunderstandings rather than acquiring desired competence with academic content, unless the teacher explicitly identifies the flaws of the game’s rules \citep{schrader2016breaking}.  

Part of the challenge in using COTS (commercial, off-the-shelf) games is that few of them are modifiable from a technical or legal standpoint. Even in games that allow modding, the modding activity is often limited by the game publisher to creating new maps for players to explore, adding customized appearances of game objects, or changing the appearance of the user interface \citep{thiel2019malleable}.  Fortunately, some commercial game developers have begun to encourage modders by the release of software development kits (SDKs) or by creating sandbox games, such as Minecraft, that are designed for active user participation and contribution of modified game components.  While these are a small portion of the gaming universe, the existence of tools that allow players to design new levels and even use provided game engines to add new features and functionality to favorite games opens up a world of opportunity for educators to leverage a popular game’s reputation and its modding community to develop educational content and distribute it to players outside of a formal learning environment  \citep{thiel2019malleable}.

However, many games that support or encourage modding, such as World of Warcraft, Call of Duty, or Half Life, are frequently focused on some sort of combat mechanic, which is  problematic when considering a game for educational use \citep{johnsonkingpanel}. Further, most of these games that can be modified do not have a world or narrative that can be appropriated to teach about the natural world and evolution via natural selection. The emergence of the `farming simulation' genre of game resolves these issues because it has little if any combat and instead focuses on the player growing crops, caring for animals, and possibly befriending non-player characters in the world. This genre is therefore friendly for educational goals and allows the incorporation of evolution via natural selection into the story narrative. A `farming simulation' game that also allows modding is Stardew Valley and therefore is the game we use for our proof-of-concept mod.

\section{Methods}

\begin{figure}[t]
  \centering
  \includegraphics[width=\linewidth]{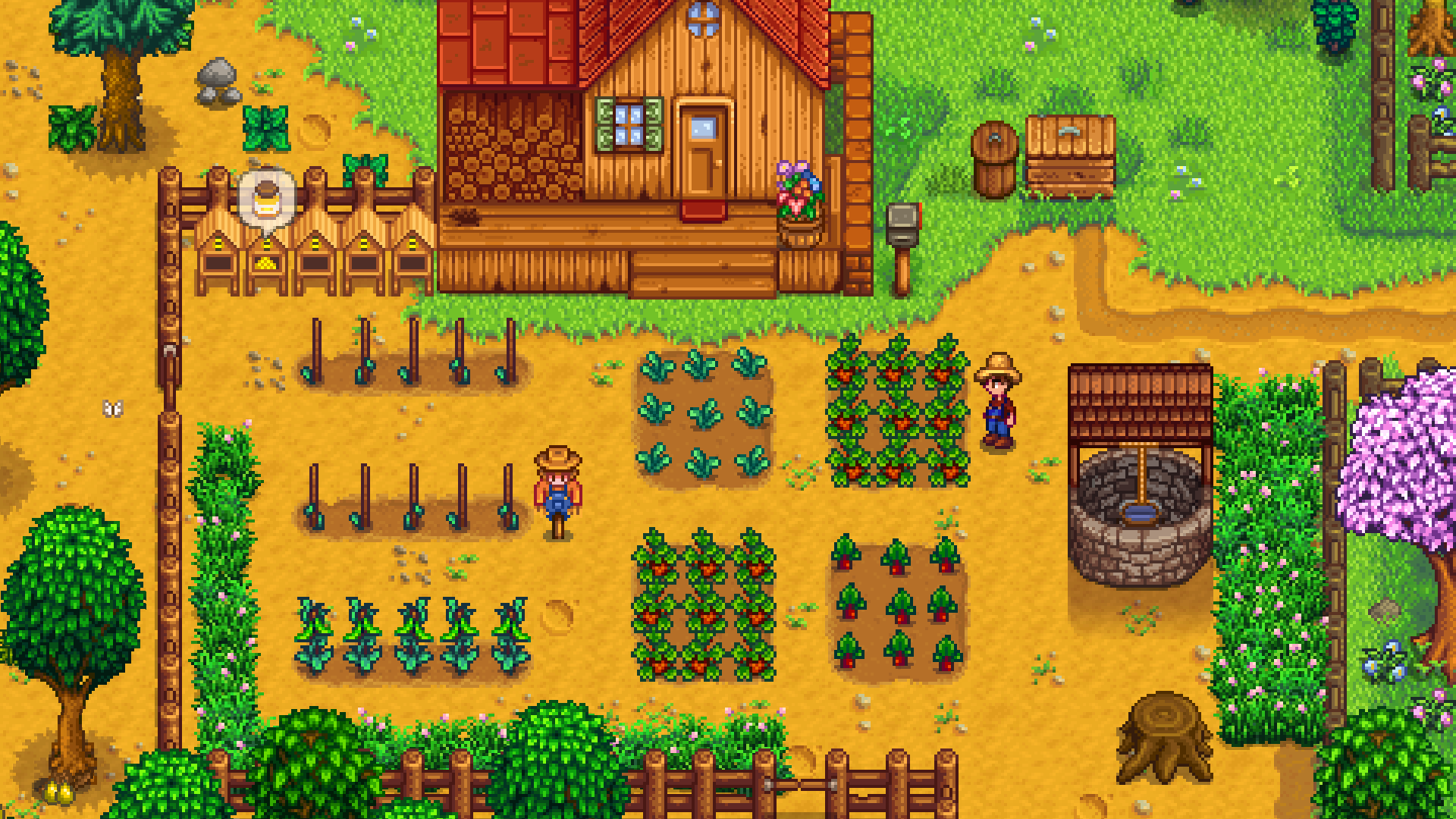}
  \caption{Screenshot from Stardew Valley of a farm with crops growing, provided by StardewValley.net.}
      \label{farm} 
\end{figure}

Stardew Valley is a semi-open world game where the player is a new arrival to a small town and has inherited  farmland that they can develop, as seen in Fig~\ref{farm}. The main activities of the game are farming crops, raising animals, fishing, mining for minerals, and interacting with the townspeople. There is arguably no ultimate goal, though earning enough money to purchase items that the player wants is a central focus. In the game there are wild and cultivated plants, livestock, wild animals, and a fishing minigame (a small game within the larger game).  All of these game mechanics are suitable places to inject artificial life, as they are modeled off of life forms already.  We focused on one of them, the fishing minigame, which allows you to catch and sell fish.

Fishing is an aspect of the game that is relatively simple and therefore developing a mod for fishing was feasible. Fishing in the base game is a dividing feature: many players enjoy fishing, so improving fishing would be desirable to them; however, many players do not enjoy fishing because the minigame is too challenging; so giving them a reason to fish would also increase the number of people using our mod.  Finally, the effects from fishing for profit is an example of rapid evolution in our world and therefore a relevant issue to demonstrate \citep{sharpe2009synthesis}.

\subsection{Fishing Basics}
In the base game, there are a finite number of species of fish that are available to be caught, and certain fish can only be caught in certain circumstances (location, season, weather, or time of day), but an infinite population of fish that is not influenced by player actions.  Which fish you catch is determined by chance, as some fish are rare and some are common.  Each fish is of a different length, in inches, but that length is used for ``fishing achievements'' when you catch the fish, and then is later discarded.  Our mod for fishing, named \textsc{RealisticFishing}, makes use of the built-in length property, but modifies it to become a part of the resulting \textsc{Item} obtained from fishing.

Two complaints often voiced about fishing in Stardew Valley are that fishing is not very profitable after the first year, and that fishing is boring.  Our mod increases player engagement with fishing by fixing both problems.  Fish are made much more profitable by adjusting the sell price to be based off of the length of the fish, rather than a fixed number.  Fishing is made more fun by pressuring the player to make informed decisions about which fish to keep, and which fish to throw back into the water.  These decisions form an intuition pattern that can be used to teach concepts about population effects and fishery collapse.

All of this is made possible by expanding the way fishing works.  There are four core changes that \textsc{RealisticFishing} makes: 1) the addition of a persistent and evolving population of fish that is affected by the player's actions, 2) the fish \textsc{Item} has been expanded to have a \textsc{length} property and that affects the sale price, 3) the player can only choose to keep 10 fish per day, and 4) the player can throw a caught fish back into the water.  These four changes produce an immediately obvious economic incentive to release small fish and keep longer fish, because they sell for more money, which then affects the persistent fish population. We want the player to think critically about which fish to keep and which fish to throw back into the water, since in the natural world, what you catch directly influences the population.  The same is true with our simulated population: catch (and keep) too many of the large fish, and your fish population is reduced to small, unprofitable fish.

We used \textsc{SMAPI}, the Stardew Valley Modding API, to create our mod.  SMAPI provides procedures to capture game events and interact with them. All of the code for \textsc{RealisticFishing} can be found on the GitHub repo here:\href{https://github.com/anyaevostinar/RealisticFishing}{https://github.com/anyaevostinar/RealisticFishing}. 

\subsubsection{General Implementation Details}

The base game does not have a persistent fish population. Instead, when the player catches a fish, there is a random chance of it being any of the species that can be found in that area. We created an actual (virtual) population of fish so that when the player catches a fish, it is randomly selected from the list of existing fish in the population. When the fish is caught (and the player chooses not to throw it back), it is removed from the population. At the beginning of each game day, the mod checks if the fish population is full and if it is not, a random fish remaining in the population reproduces (asexually) with a chance of its offspring having a mutated length trait. The amount of mutation is pulled from a normal distribution with a mean of 0 and a standard deviation of 2, which is then added to the parent’s length trait value to determine the offspring’s length trait value. There is a maximum and minimum length allowed for each type of fish, and so if the offspring exceeds these limits, its length is set to the exceeded bound.

While a length trait was pre-existing in the game, it did not actually factor into the price that the player could sell the fish for, nor was it tracked by the game after the player caught the fish. To enable there to be some reason for the player to care about the length of the fish they catch, we added the length trait to the fish \textsc{Item}. The amount of money the player receives when they sell a fish is now whatever the fish would have sold for in the base game plus the length of the fish divided by 8. We decided on the divisor of 8 through playtesting and determining what seemed balanced.

With the previous two changes, the player still would likely not actually see evolution in the fish population because the base game allows the player to keep all the fish that they catch. To force the player to make decisions about which fish to keep, and to improve the realism of the mod, we introduced a daily limit of 10 fish. This addition is similar to many fishing restrictions in the real world \citep{heino2002fisheries} and causes there to be an incentive to the player only keeping fish that are the most valuable, which are the fish that are the longest. To enable the player to keep only the longer fish, we also added the ability for the player to throw back fish that they do not want to keep. 

These changes mean that a player that aims to make as much money as possible will keep only the ten largest fish that they catch each day. This decision will remove those fish from the population, leaving fish that are smaller on average to reproduce. If this decision is repeated, the fish population will evolve to be smaller on average, as has occurred in some real world fisheries \citep{sharpe2009synthesis}. 

\begin{figure}[t]
  \centering
  \includegraphics[width=\linewidth]{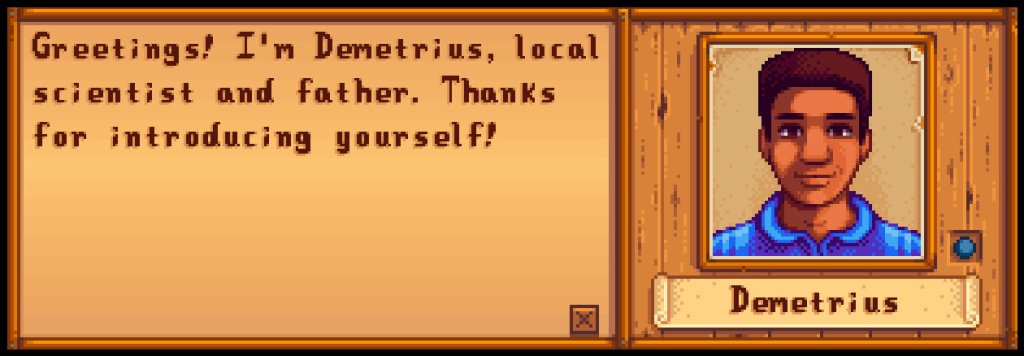}
  \caption{Screenshot from Stardew Valley of the dialogue when Demetrius, the resident scientist, introduces himself.}
      \label{demetrius} 
\end{figure}

While the player may notice this change in their fish population on their own, we decided to make it especially clear to them by expanding another existing game feature. The game includes a scientist in the town already, shown in Fig~\ref{demetrius}, (an inclusion that the authors find wonderful) and the player can receive mail from the townspeople through normal game play. We added the feature that if the average length of a fish type in the population is below a threshold, the scientist sends the player a letter stating:
\begin{quote}

``Dear \textsc{player}, I was conducting a field study on\\ \textsc{fishName} the other day, and I discovered that the population is in decline. To prevent a fishery collapse, please release any large \textsc{fishName} you catch until the population is stable again. -Demetrius''

\end{quote}

\section{Results and Discussion}
The goal of this work has been to evaluate and present the feasibility of modifying an existing popular game to incorporate evolutionary algorithms and artificial life principles for educational purposes. We note that we have not performed educational studies to determine if and how much this mod changes understanding and acceptance of evolution, but are pursuing that work currently. We present this mod as a proof of concept only at this point.

We were able to use an evolutionary algorithm to create evolutionary dynamics in the game Stardew Valley through the fishing feature of the game. The mod is available for players to download at \href{https://www.nexusmods.com/stardewvalley/mods/2623}{https://www.nexusmods.com/stardewvalley/mods/2623}. As of February 12th, 2020 the mod has been viewed 10,265 times, downloaded 244 times, and there have been several interesting comments on the mod page, such as ``This mod is realistic, yes you could fish more than 10 fish in reality, but it teaches you not to fish too much and about sustainable fishing'' \citep{nexuscomment}.

Through this process, we have determined that Stardew Valley may not actually be the ideal platform for this kind of work for several reasons. First, the code base of Stardew Valley is difficult to modify. While there are tools to help and a vibrant community, the existing codebase is structured in such a way that some things are currently impossible to modify. For example, it would be desirable to introduce evolution of the farm animals, which are a more central aspect of the game. Unfortunately, any modification to the farm animals beyond aesthetics is not currently possible. In addition, Stardew Valley, like most modern games, is continuously being updated and the modding tool SMAPI is updated along with the main game. This would not be a problem if older versions of both were easily available, but they are not, and therefore a mod would need to be continuously updated to continue to be used. 

Our preliminary results indicate that the strategy of modding existing games to incorporate evolutionary algorithms is worth further exploration, however there may be other platforms that are easier to employ this strategy with. Another popular game, Minecraft, supports modification and also has an extensive modding community. Minecraft avoids the issue of continuous updates by continuing to support older versions of the game and many popular mods only work on older versions, but players are willing to stick to that older version to use the mod. Minecraft also has animals, crops, and fishing which can all be modified extensively.

\section{Conclusion}
We have shown that it is possible to implement evolutionary dynamics in an existing popular game and deploy that modification to the general public. In future work, we will conduct experiments to determine if and how much such a modification increases players’ understanding and acceptance of evolution. In addition, we will test the feasibility and reception of similar modifications to other popular and modifiable games. 

By modifying pre-existing games, we are able to place accurate evolutionary dynamics into games that are already fun and widely used. Further, the way that the evolutionary dynamics are incorporated can be chosen to be subtle enough that players do not realize that they are witnessing evolution in action, possibly making some more receptive to understanding the dynamics of evolution, as has been done with other forms of stealth gaming. While this strategy of modification is only one of many ways to increase evolution understanding and acceptance, it is one that may reach further and therefore is worth considering.

\section{Acknowledgements}

This project was funded by the Mentored Advanced Project program at Grinnell College. The authors would also like to thank Eric Barone for creating Stardew Valley and the modder Pathoschild who created SMAPI and has always been willing to answer questions as we developed RealisticFishing.

\footnotesize
\bibliographystyle{apalike}
\balance
\bibliography{example} 

\begin{thebibliography}{}

\bibitem[Baker, 2008]{baker2008trying}
Baker, C. (2008).
\newblock Trying to design a truly entertaining game can defeat even a
  certified genius.
\newblock {\em Wired Magazine}, 16(04).

\bibitem[Barab et~al., 2005]{barab2005making}
Barab, S., Thomas, M., Dodge, T., Carteaux, R., and Tuzun, H. (2005).
\newblock Making learning fun: Quest atlantis, a game without guns.
\newblock {\em Educational technology research and development}, 53(1):86--107.

\bibitem[Barone, 2020]{stardew_valley}
Barone, E. (2020).
\newblock {Stardew Valley Homepage}.

\bibitem[Bean et~al., 2010]{bean2010spore}
Bean, T.~E., Sinatra, G.~M., and Schrader, P. (2010).
\newblock Spore: Spawning evolutionary misconceptions?
\newblock {\em Journal of Science Education and Technology}, 19(5):409--414.

\bibitem[Bertozzi, 2014]{bertozzi2014using}
Bertozzi, E. (2014).
\newblock Using games to teach, practice, and encourage interest in stem
  subjects.
\newblock {\em Learning, Education and Games}, page~23.

\bibitem[Christensen-Dalsgaard and Kanneworff, 2009]{christensen2009evolution}
Christensen-Dalsgaard, J. and Kanneworff, M. (2009).
\newblock Evolution in lego{\textregistered}: a physical simulation of
  adaptation by natural selection.
\newblock {\em Evolution: Education and Outreach}, 2(3):518--526.

\bibitem[Clarke and Dede, 2009]{clarke2009design}
Clarke, J. and Dede, C. (2009).
\newblock Design for scalability: A case study of the river city curriculum.
\newblock {\em Journal of Science Education and Technology}, 18(4):353--365.

\bibitem[Dahleen et~al., 2014]{dahleen2014towards}
Dahleen, J., Hunsberger, A., Weber, R., Brylow, D., Longstreet, C.~S., and
  Cooper, K.~M. (2014).
\newblock Towards a lightweight approach for modding serious educational games:
  Assisting novice designers.
\newblock In {\em Proceedings of the 20th International Conference on
  Distributed Multimedia Systems}. Knowledge Systems Institute.

\bibitem[Gaydos and Squire, 2012]{gaydos2012role}
Gaydos, M.~J. and Squire, K.~D. (2012).
\newblock Role playing games for scientific citizenship.
\newblock {\em Cultural Studies of Science Education}, 7(4):821--844.

\bibitem[Gibson et~al., 2015]{gibson2015red}
Gibson, A.~K., Drown, D.~M., and Lively, C.~M. (2015).
\newblock The red queen’s race: An experimental card game to teach
  coevolution.
\newblock {\em Evolution: Education and Outreach}, 8(1):10.

\bibitem[Hayes and King, 2009]{hayes2009not}
Hayes, E.~R. and King, E.~M. (2009).
\newblock Not just a dollhouse: what the sims2 can teach us about women's it
  learning.
\newblock {\em On The Horizon-The Strategic Planning Resource for Education
  Professionals}, 17(1):60--69.

\bibitem[Heino and God{\o}, 2002]{heino2002fisheries}
Heino, M. and God{\o}, O.~R. (2002).
\newblock Fisheries-induced selection pressures in the context of sustainable
  fisheries.
\newblock {\em Bulletin of marine science}, 70(2):639--656.

\bibitem[Johnson et~al., 2008]{johnsonkingpanel}
Johnson, B.~Z., King, E., and Yucel, I. (2008).
\newblock User-created content and program-modification in video games and
  virtual worlds.
\newblock Panel at Meaningful Play.

\bibitem[J{\"o}rdens et~al., 2018]{jordens2018interrelating}
J{\"o}rdens, J., Asshoff, R., Kullmann, H., and Hammann, M. (2018).
\newblock Interrelating concepts from genetics and evolution: why are cod
  shrinking?
\newblock {\em The American Biology Teacher}, 80(2):132--138.

\bibitem[Krathwohl and Anderson, 2009]{krathwohl2009taxonomy}
Krathwohl, D.~R. and Anderson, L.~W. (2009).
\newblock {\em A taxonomy for learning, teaching, and assessing: A revision of
  Bloom's taxonomy of educational objectives}.
\newblock Longman.

\bibitem[Lave et~al., 1991]{lave1991situated}
Lave, J., Wenger, E., et~al. (1991).
\newblock {\em Situated learning: Legitimate peripheral participation}.
\newblock Cambridge university press.

\bibitem[Monterrat et~al., 2012]{monterrat2012learning}
Monterrat, B., Lavou{\'e}, E., and George, S. (2012).
\newblock Learning game 2.0: Support for game modding as a learning activity.
\newblock In {\em Proceedings of the 6th European Conference on Games Based
  Learning}, pages 340--347.

\bibitem[Moshirnia, 2007]{moshirnia2007educational}
Moshirnia, A. (2007).
\newblock The educational potential of modified video games.
\newblock {\em Issues in Informing Science \& Information Technology}, 4.

\bibitem[OolongBoolong, 2020]{nexuscomment}
OolongBoolong (2020).
\newblock Realistic fishing mod page comment.
\newblock
  \url{https://forums.nexusmods.com/index.php?/topic/6885712-realisticfishing/page-2#entry75701083}.

\bibitem[Orland, 2018]{valve_leak}
Orland, K. (2018).
\newblock {Valve leaks Steam game player counts; we have the numbers}.

\bibitem[Prensky, 2010]{prensky2010teaching}
Prensky, M.~R. (2010).
\newblock {\em Teaching digital natives: Partnering for real learning}.
\newblock Corwin press.

\bibitem[Schrader et~al., 2016]{schrader2016breaking}
Schrader, P., Deniz, H., and Keilty, J. (2016).
\newblock Breaking spore: Building instructional value in science education
  using a commercial, off-the shelf game.
\newblock {\em Journal of Learning and Teaching in Digital Age}, 1(1):63--73.

\bibitem[Sharp, 2012]{sharp2012stealth}
Sharp, L.~A. (2012).
\newblock Stealth learning: Unexpected learning opportunities through games.
\newblock {\em Journal of Instructional Research}, 1:42--48.

\bibitem[Sharpe and Hendry, 2009]{sharpe2009synthesis}
Sharpe, D.~M. and Hendry, A.~P. (2009).
\newblock Synthesis: life history change in commercially exploited fish stocks:
  an analysis of trends across studies.
\newblock {\em Evolutionary applications}, 2(3):260--275.

\bibitem[Shreve, 2005]{shreve2005let}
Shreve, J. (2005).
\newblock Let the games begin. video games, once confiscated in class, are now
  a key teaching tool. if they're done right.
\newblock {\em George Lucas Educational Foundation}.

\bibitem[Sinatra et~al., 2008]{sinatra2008changing}
Sinatra, G.~M., Brem, S.~K., and Evans, E.~M. (2008).
\newblock Changing minds? implications of conceptual change for teaching and
  learning about biological evolution.
\newblock {\em Evolution: Education and outreach}, 1(2):189--195.

\bibitem[Squire and Barab, 2004]{squire2004replaying}
Squire, K. and Barab, S. (2004).
\newblock Replaying history: Engaging urban underserved students in learning
  world history through computer simulation games.
\newblock In {\em Proceedings of the 6th international conference on Learning
  sciences}, pages 505--512. International Society of the Learning Sciences.

\bibitem[Steinkuehler and Johnson, 2009]{steinkuehler2009computational}
Steinkuehler, C.~C. and Johnson, B.~Z. (2009).
\newblock Computational literacy in online games: The social life of mods.
\newblock {\em International Journal of Gaming and Computer-Mediated
  Simulations (IJGCMS)}, 1(1):53--65.

\bibitem[Thiel and Lyle, 2019]{thiel2019malleable}
Thiel, S.-K. and Lyle, P. (2019).
\newblock Malleable games-a literature review on communities of game modders.
\newblock In {\em Proceedings of the 9th International Conference on
  Communities \& Technologies-Transforming Communities}, pages 198--209.

\end{thebibliography}

\end{document}